\documentclass[11pt]{article}

\usepackage[final]{acl}

\usepackage{times}
\usepackage{latexsym}
\usepackage{amsmath}

\usepackage{times}
\usepackage{latexsym}
\usepackage[T1]{fontenc}
\usepackage[utf8]{inputenc}
\usepackage{microtype}
\usepackage{inconsolata}
\usepackage{graphicx}
\usepackage{booktabs}
\usepackage{multirow}
\usepackage{amsmath}
\usepackage{amssymb}
\usepackage{algorithm}
\usepackage{algpseudocode}
\usepackage{url}
\usepackage{enumitem}
\usepackage{xcolor}
\usepackage[table]{xcolor}
\usepackage{subcaption}
\definecolor{tableaublue}{HTML}{1f77b4}
\usepackage{hyperref}

\usepackage[T1]{fontenc}

\usepackage[utf8]{inputenc}

\usepackage{microtype}

\usepackage{inconsolata}

\usepackage{graphicx}
\newcommand{\method}{\textsc{Scaffold}\xspace}
\usepackage{xspace}

\newcommand*{\img}[1]{%
    \raisebox{-.01\baselineskip}{%
        \includegraphics[
        height=2\baselineskip,
        width=2\baselineskip,
        keepaspectratio,
        ]{#1}%
    }%
}

\usepackage{listings}
\definecolor{diffadd}{HTML}{22863A}
\definecolor{diffrem}{HTML}{B31D28}
\definecolor{diffhunk}{HTML}{6A737D}
\definecolor{diffbg}{HTML}{F6F8FA}
 
\lstdefinelanguage{gitdiff}{
  basicstyle=\ttfamily\scriptsize,
  backgroundcolor=\color{diffbg},
  numbers=none,
  showspaces=false,
  showstringspaces=false,
  showtabs=false,
  breaklines=true,
  breakatwhitespace=true,
  columns=fullflexible,
  keepspaces=true,
  frame=single,
  framerule=0pt,
  framesep=3pt,
  xleftmargin=4pt,
  xrightmargin=4pt,
  morecomment=[l][\color{diffrem}]{-},
  morecomment=[l][\color{diffadd}]{+},
  morecomment=[l][\color{diffhunk}\itshape]{@@},
}

\title{\img{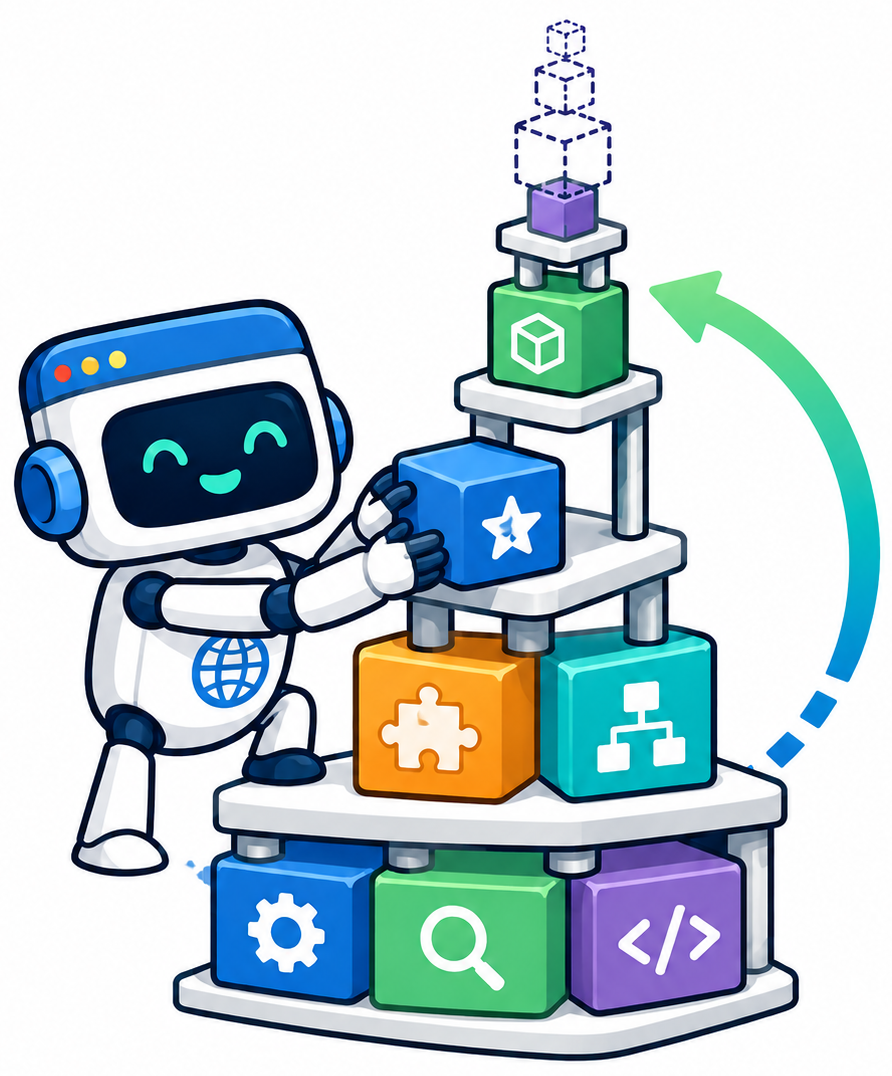} SCAFFOLD: Self-Improving Web Agents via Recursive \\ Parametric Skill Abstraction}

\author{Bowei He\textsuperscript{1, 2}, Xiaokun Zhang\textsuperscript{3}, Meng Ding\textsuperscript{4}, Xue Liu\textsuperscript{1, 2} \\
\textsuperscript{1} MBZUAI, \textsuperscript{2} McGill University, \textsuperscript{3} CityUHK,
\textsuperscript{4} UMass Boston \\
  \texttt{Bowei.He@mbzuai.ac.ae}}

\begin{document}
\maketitle
\begin{abstract}
Web agents need to navigate visually rich, long-horizon interfaces that change across sites, yet most previous agents still learn each task in isolation and discard the procedural knowledge they accumulate. Recent skill-augmented frameworks take an important first step, but they treat the skill library as a flat or two-tier prompt-side cache and offer no principled mechanism for compressing redundancy or composing skills recursively. We introduce \textsc{Scaffold}, a self-improving framework for visual web agents that (i) induces parametric, executable skills from successful trajectories under a multi-instance abstraction constraint, (ii) maintains a recursively composed hierarchy in which higher-level skills invoke lower-level ones, (iii) compacts the library via a minimum-description-length (MDL) criterion and behavioral equivalence checking, and (iv) periodically distills skill-augmented trajectories back into model weights to internalize the abstractions. Across WebArena, VisualWebArena, and a held-out split of Online-Mind2Web, \textsc{Scaffold} improves success rate by $11.1$--$17.2$ absolute points over the strongest skill-augmented baseline and shows monotonic gains across five self-improvement iterations without library collapse. We release the code and documents in the Github \href{https://github.com/BokwaiHo/SCAFFOLD}{repository}.
\end{abstract}

\section{Introduction}
\label{sec:intro}
Autonomous web agents, LLM-based systems that complete user instructions by interacting with browsers, have advanced rapidly with the arrival of strong multimodal foundation models~\cite{zheng2024seeact,he2024webvoyager,qin2025uitars,seed2025uitars2}. Yet on every realistic benchmark, from WebArena~\cite{zhou2024webarena} and VisualWebArena~\cite{koh2024visualwebarena} to OSWorld~\cite{xie2024osworld} and Online-Mind2Web~\cite{xue2025illusion}, the performance still lags human performance by a wide margin. A central cause is that current agents treat each task as an isolated episode: procedural knowledge acquired while booking a flight is discarded before the next task begins, even if that next task shares most of the same sub-procedure~\cite{wang2024agentworkflow,zheng2025skillweaver}.
 
This observation has motivated a wave of \textit{self-improving} web agents that retain experience in some form. Three sub-paradigms have emerged. \textbf{Trajectory replay} methods store raw experience for retrieval-augmented prompting~\cite{zheng2024synapse,zhao2024expel,chhikara2025mem0}. \textbf{Workflow induction} methods abstract action sequences into natural-language routines that the agent can re-read at inference time~\cite{wang2024agentworkflow,fang2026trajectory}. \textbf{Skill induction} methods go further and synthesize callable APIs or parametric skills, exemplified by \textsc{SkillWeaver}~\cite{zheng2025skillweaver}, \textsc{AppAgentX}~\cite{jiang2025appagentx}, and the very recent \textsc{SkillRL}~\cite{xia2026skillrl}. Skill induction is the strongest of them because the reuse unit is an executable program rather than a textual hint, and skills can in principle be \textit{composed}. This property, as long argued by the hierarchical reinforcement learning community~\cite{sutton1999options}, is essential for tackling long-horizon problems.
 
Despite this trajectory, three limitations of current skill-augmented agents remain unresolved.\footnote{We focus on visual web agents throughout; complementary work on text-only LLM agents includes~\cite{xia2026skillrl,wang2026reinforcement,wu2025evolver}.} First, induced skill libraries are essentially flat: \textsc{SkillWeaver} stores APIs in a single pool, and \textsc{SkillRL} uses only a two-tier (general / task-specific) split. Since neither supports skills calling other skills, the maximum abstraction depth is bounded by one. Second, libraries grow monotonically and accumulate near-duplicate skills with no principled compression mechanism, a known failure mode in lifelong learning~\cite{wang2023voyager}. Third, the knowledge encoded in skills exists only in prompts and is never internalized into the model, leaving the base policy after iterations as weak as it was at the beginning.
 
\vspace{0.3em}
\noindent\textbf{Contributions.} We introduce \method, a framework that addresses each of these limitations:
\begin{itemize}[leftmargin=*, itemsep=1pt, topsep=2pt]
\item A \textit{multi-instance parameter induction} procedure that proposes a skill only when multiple semantically similar trajectories support the same parametric abstraction, sharply reducing spurious abstracted skills (§\ref{sec:induction}).
\item A \textit{recursive composition} mechanism: at iteration $k$, the inducer is allowed to call any skill from iterations $1,\ldots,k$, producing a hierarchy whose depth grows monotonically. We track depth and reuse explicitly in our design (§\ref{sec:composition}).
\item An \textit{MDL-driven library compaction} step that periodically merges behaviorally equivalent skills, refactors recurring sub-patterns into new mid-level skills, and prunes unused ones, with empirical validation on a held-out task set (§\ref{sec:compaction}).
\item A \textit{distillation loop} that uses skill-augmented trajectories to supervise-fine-tune the base policy, converting the in-context library into improved weights (§\ref{sec:distillation}).
\end{itemize}
 
On WebArena, VisualWebArena, and a held-out Online-Mind2Web subset, \method improves over \textsc{SkillWeaver} by 13.6--17.7 absolute success-rate points. Critically, it continues to improve through five self-improvement iterations (whereas baselines saturate at two), exhibits a desirable library-depth and reuse-rate distribution, and zero-shot transfers to held-out sites with a 17.2-point margin over the strongest competitor.
 
\section{Related Work}
\label{sec:related}
\noindent\textbf{Web and GUI agents.}
Modern web agents either prompt strong proprietary models~\cite{zheng2024seeact,he2024webvoyager,yang2025agentoccam} or fine-tune open vision-language models on large GUI trajectory corpora~\cite{cheng2024seeclick,hong2024cogagent,wu2025atlas,xu2025aguvis,qin2025uitars,seed2025uitars2,lai2024autowebglm}. The latter line achieves impressive grounding but relies on static datasets that fail to capture the procedural diversity of real websites~\cite{zheng2025skillweaver}. Reinforcement-learning-based agents close part of this gap by training on the environment directly~\cite{qi2025webrl,patel2024large,gandhi2026gobrowse,shao2024deepseekmath,guo2025deepseekr1}, but typically lack any mechanism for accumulating reusable behavioral abstractions, which is our focus.
 
\noindent\textbf{Self-improvement through experience.}
The idea of letting a model improve itself by training on its own filtered generations dates back to \textsc{STaR}~\cite{zelikman2022star} and was extended to richer settings by Quiet-STaR~\cite{zelikman2024quietstar} and self-play fine-tuning~\cite{chen2024self}. For agents specifically, \cite{patel2024large} showed that filtered self-generated trajectories can lift WebArena performance, and Reflexion~\cite{shinn2023reflexion} introduced verbal in-context self-correction without weight updates. \method inherits the spirit of these works but operates over a \textit{structured} skill library rather than raw text or trajectory data.
 
\noindent\textbf{Skill libraries and workflows.}
\textsc{Voyager}~\cite{wang2023voyager} pioneered LLM-driven skill libraries in the programmatic Minecraft environment; \textsc{TroVE}~\cite{wang2024trove} extended this idea to tool induction for programmatic tasks. For web/GUI agents, Agent Workflow Memory (AWM)~\cite{wang2024agentworkflow} induces natural-language workflows from past trajectories and retrieves them at test time, achieving large gains on Mind2Web and WebArena. \textsc{SkillWeaver}~\cite{zheng2025skillweaver} advances this paradigm by synthesizing skills as executable APIs and using iterative practice for refinement. \textsc{AppAgentX}~\cite{jiang2025appagentx} explores analogous ideas on mobile UIs. The most direct prior work is \textsc{SkillRL}~\cite{xia2026skillrl}, which trains an LLM agent jointly with a two-tier hierarchical skill library (general skills + task-specific skills) using GRPO~\cite{shao2024deepseekmath}. \method differs in operating on visual web agents with truly recursive hierarchies, an MDL-based compaction step, and an explicit skill transfer evaluation.

\noindent\textbf{Concurrent skill frameworks.}
\textsc{PolySkill}~\cite{yu2026polyskill} separates a skill's abstract goal from its per-site implementations via abstract interfaces, whereas our skills re-bind semantic references at run time; \textsc{SkillEvo}~\cite{skillevo2026} pairs GRPO with an evolving skill path graph whose reuse unit is an experience path rather than a parametric program and which carries no compression objective. 
 
\noindent\textbf{Memory and lifelong learning for agents.}
A parallel literature treats agent experience as memory rather than skill, including \textsc{Mem0}~\cite{chhikara2025mem0}, \textsc{ExpeL}~\cite{zhao2024expel}, \textsc{EvolveR}~\cite{wu2025evolver}, \textsc{Synapse}~\cite{zheng2024synapse}, and trajectory-informed memory generation~\cite{fang2026trajectory}. These methods are largely orthogonal to skill induction and could be combined with \method; we use \textsc{Mem0} and \textsc{Synapse} as memory-style baselines in §\ref{sec:experiments}. Finally, classical hierarchical RL~\cite{sutton1999options} and library learning in program synthesis~\cite{ellis2021dreamcoder} provide theoretical grounding for the recursive composition and MDL-based compression operations central to our method.
 
\begin{figure*}[t]
\centering
\includegraphics[width=0.98\textwidth]{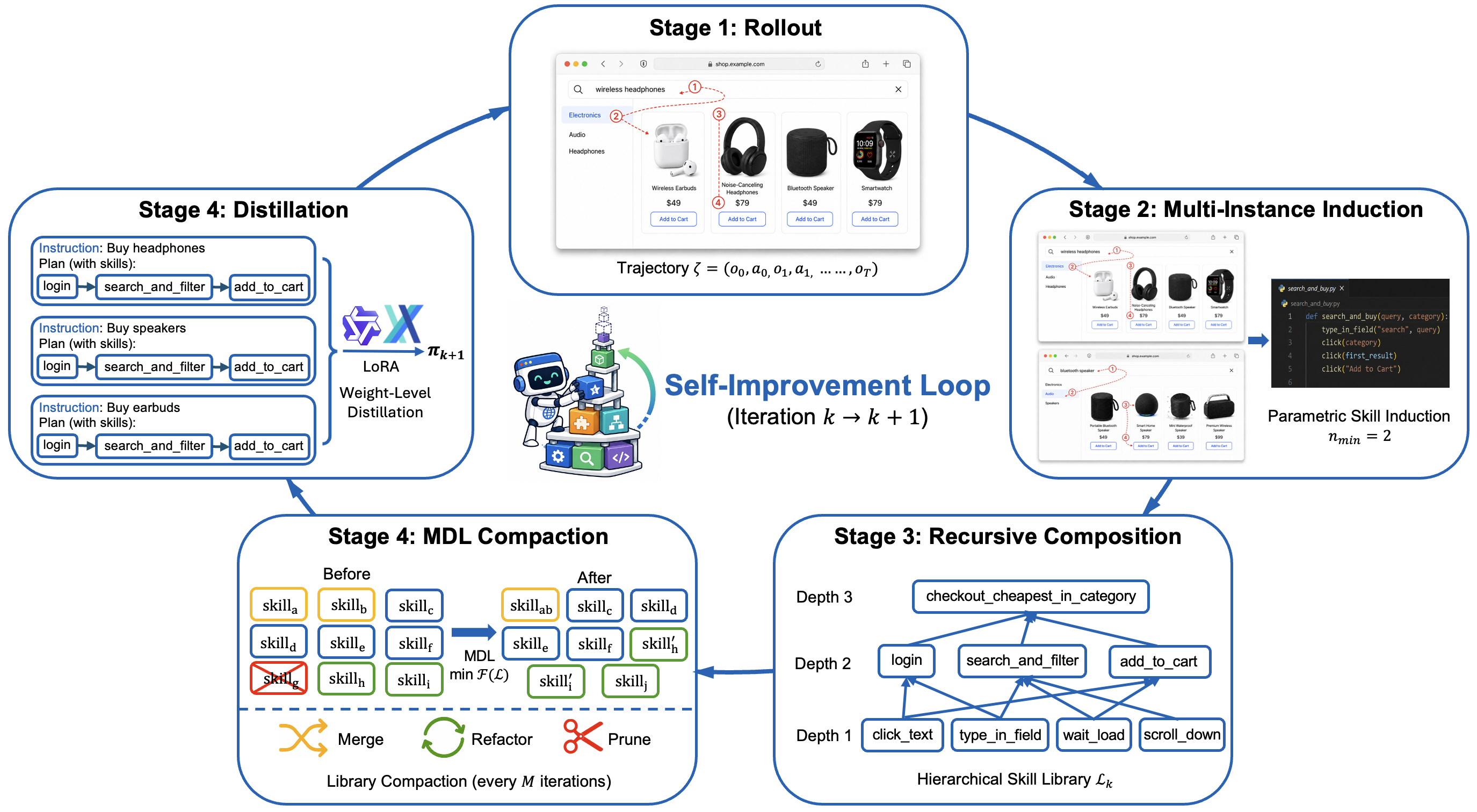}
\vspace{-1mm}
\caption{Overview of \method framework. The five stages are executed sequentially in each iteration: rollout, multi-instance induction, recursive composition, MDL-driven compaction, and distillation. The library growing in both depth and width is periodically compressed.}
\vspace{-3mm}
\label{fig:framework}
\end{figure*}

\section{Methodology}
\label{sec:method}
\subsection{Problem Setup and Notation}
\label{sec:setup}
We consider an agent operating in a partially observable web environment $\mathcal{E}$. At step $t$, the agent receives observation $o_t = (s_t, d_t, u_t)$ consisting of a screenshot $s_t$, accessibility-tree / document object model (DOM) snippet $d_t$, and current URL $u_t$. The agent emits an action $a_t \in \mathcal{A}_{\text{prim}} \cup \mathcal{A}_{\text{skill}}$, where $\mathcal{A}_{\text{prim}}$ are primitive actions (click, type, scroll, wait) and $\mathcal{A}_{\text{skill}}$ is the set of currently available skill invocations. A task $\tau = (I, \text{init}, V)$ specifies a natural-language instruction $I$, an initial environment state, and a verifier $V$ that returns binary success on the final state. We denote a trajectory by $\zeta = (o_0, a_0, o_1, \ldots, o_T)$.
A \textbf{skill} is a tuple $\sigma = (\text{name}, \text{description}, \boldsymbol{\theta}, \text{pre}, \text{body}, \text{post}, d_\sigma)$. $\boldsymbol{\theta}$ is a typed parameter list admitting primitive values (strings, numbers), semantic element references (natural-language descriptors grounded to DOM elements at run time by a VLM grounder), and pointers to other skills. $\text{pre}$ and $\text{post}$ are LLM-checkable preconditions and postconditions phrased as predicates over $o_t$, $\text{body}$ is an executable program over $\mathcal{A}_{\text{prim}} \cup \mathcal{A}_{\text{skill}}$, and $d_\sigma \in \mathbb{N}$ is the \textit{abstraction depth}. The library $\mathcal{L}_k$ at iteration $k$ is a set of such skills together with an embedding-based retrieval index keyed on $\text{description}$. We call $\sigma$ \textit{semi-parametric}: control flow inside $\text{body}$ calls no policy model, while semantic element references are grounded on the live observation by a small VLM (§\ref{sec:budget}).
 
\subsection{Overall Pipeline}
\label{sec:pipeline}
Each self-improvement iteration $k \to k+1$ consists of five stages, executed in order: (1) \textit{trajectory collection} with the current policy $\pi_k$ and library $\mathcal{L}_k$; (2) \textit{multi-instance skill induction}; (3) \textit{recursive composition} of new skills over $\mathcal{L}_k$; (4) \textit{MDL-driven library compaction}; and (5) \textit{distillation} of the skill-augmented behavior into $\pi_{k+1}$. Algorithm~\ref{alg:scaffold} summarizes the loop. Below we describe each stage in detail; the design choices are motivated by the failure modes of prior skill-induction methods that we observe empirically (cf. §\ref{sec:ablation}). The overall framework is also illustrated in Figure~\ref{fig:framework}.
\vspace{-1mm}

\subsection{Multi-Instance Skill Induction}
\label{sec:induction}
A persistent failure mode in workflow-induction methods is the synthesis of \textit{spurious abstractions}: a skill is proposed from a single trajectory but encodes incidental details (specific element IDs, hard-coded text) that fail to generalize. We address this by requiring every candidate skill to be supported by at least $n_{\min}$ trajectories with the same parameter structure.
 
Successful trajectories $\mathcal{Z}_k^+$ are first clustered by the embedding of their instruction $I$. Within each cluster $C = \{\zeta_1, \ldots, \zeta_n\}$ with $n \geq n_{\min}$, the inducer (a frozen LLM) is prompted with all $n$ trajectories and asked to: (a) identify which positions vary across trajectories and propose a typed parameter list $\boldsymbol{\theta}$, (b) emit an executable body using $\boldsymbol{\theta}$, primitives in $\mathcal{A}_{\text{prim}}$, and any skill in $\mathcal{L}_k$, (c) state a precondition / postcondition pair as Python expressions over $o_t$. The induced skill must \textit{re-execute correctly} on at least one held-out instance from $C$; otherwise it is discarded. This validation step replaces the post-hoc filtering used in \textsc{SkillWeaver} and AWM, and we show in §\ref{sec:ablation} that it nearly halves the rate of induced-then-deprecated skills across iterations.

\begin{algorithm}[t]
\small
\caption{\method Self-Improvement Loop}
\label{alg:scaffold}
\begin{algorithmic}[1]
\Require Base policy $\pi_0$, task pool $\mathcal{T}$, verifier $V$, iterations $K$, compaction interval $M$
\State $\mathcal{L}_0 \gets \emptyset$
\For{$k = 0, \ldots, K-1$}
  \State $\mathcal{Z}_k \gets \textsc{Rollout}(\pi_k, \mathcal{L}_k, \mathcal{T})$ \Comment{collect trajectories}
  \State $\mathcal{Z}_k^+ \gets \{\zeta \in \mathcal{Z}_k : V(\zeta)=1\}$
  \State $\mathcal{C}_k \gets \textsc{ClusterByInstr}(\mathcal{Z}_k^+)$ \Comment{group similar trajs}
  \State $\Sigma^{\text{new}} \gets \emptyset$
  \For{cluster $C \in \mathcal{C}_k$ with $|C| \geq n_{\min}$}
    \State $\sigma \gets \textsc{Induce}(C, \mathcal{L}_k)$ \Comment{multi-instance abstraction; may invoke any $\sigma' \in \mathcal{L}_k$}
    \If{$\textsc{ValidateHoldout}(\sigma, \mathcal{T}_{\text{val}})$}
      \State $\Sigma^{\text{new}} \gets \Sigma^{\text{new}} \cup \{\sigma\}$
    \EndIf
  \EndFor
  \State $\mathcal{L}_{k+1} \gets \mathcal{L}_k \cup \Sigma^{\text{new}}$
  \If{$(k+1) \bmod M = 0$}
    \State $\mathcal{L}_{k+1} \gets \textsc{Compact}_{\text{MDL}}(\mathcal{L}_{k+1}, \mathcal{Z}_{0:k}^+)$
  \EndIf
  \State $\pi_{k+1} \gets \textsc{Distill}(\pi_k, \mathcal{L}_{k+1}, \mathcal{Z}_k^+)$
\EndFor
\State \Return $\pi_K, \mathcal{L}_K$
\end{algorithmic}
\end{algorithm}
 
\subsection{Recursive Composition and Depth Tracking}
\label{sec:composition}
The library is hierarchical by construction: when the inducer emits a body, it may invoke any skill $\sigma' \in \mathcal{L}_k$ as a sub-routine. The depth of a new skill $\sigma$ is defined as
\begin{equation}
d_\sigma = 1 + \max_{\sigma' \in \text{calls}(\sigma)} d_{\sigma'},
\label{eq:depth}
\end{equation}
with $d_\sigma = 1$ when $\text{calls}(\sigma) \subseteq \mathcal{A}_{\text{prim}}$. We do not enforce a maximum depth; instead we measure how depth evolves across iterations (§\ref{sec:depth}). To prevent runaway abstraction, we forbid cycles via a simple topological check at induction time, and we cap the per-skill body length at 30 lines.
 
The choice to allow \textit{any} prior skill, not just skills from iteration $k-1$, enables the inducer to refactor a long chain of primitives into a single mid-level skill at any iteration. Concretely, an induced \texttt{book\_flight} skill at iteration $3$ may call \texttt{search\_dates} (iteration $1$), \texttt{fill\_passenger\_info} (iteration $2$), and a freshly synthesized \texttt{confirm\_payment} (also iteration $3$). This stands in contrast to \textsc{SkillRL}'s fixed two-tier hierarchy, which forces the agent to flatten the natural call structure of real web procedures.
 
\subsection{MDL-Driven Library Compaction}
\label{sec:compaction}
Without governance, the library grows monotonically and accumulates redundant skills, which both inflates the retrieval namespace and dilutes the gradient signal during distillation. Every $M$ iterations (we use $M{=}2$) we apply a \textit{compaction} step whose objective is the classical minimum-description-length functional~\cite{rissanen1978mdl}:
\begin{equation}
\mathcal{F}(\mathcal{L}) = \underbrace{\sum_{\sigma \in \mathcal{L}} |\sigma|}_{\text{library cost}} + \underbrace{\sum_{\zeta \in \mathcal{Z}^+} \min_{\text{parse}(\zeta\mid\mathcal{L})} |{\text{parse}(\zeta\mid\mathcal{L})}|}_{\text{data cost}}.
\label{eq:mdl}
\end{equation}
Here $|\sigma|$ is the token length of the skill body, and $|\text{parse}(\zeta \mid \mathcal{L})|$ is the length of the shortest program over $\mathcal{L} \cup \mathcal{A}_{\text{prim}}$ that reproduces $\zeta$. Intuitively, $\mathcal{F}$ rewards libraries that are short (few, concise skills) but that can compactly explain past successful behavior; it penalizes both \textit{over-specific} skills (used by few trajectories) and \textit{redundant} skills (whose role is already covered).
 
We approximate the minimization of Eq.~\ref{eq:mdl} via three greedy operators applied iteratively until $\mathcal{F}$ no longer decreases:
\textbf{(i) Merge}: two skills $\sigma_a, \sigma_b$ that are \textit{behaviorally equivalent} on a sampled holdout set (i.e., produce identical post-states with probability $\geq \rho$) are merged into the shorter of the two, with the longer being replaced by an alias.
\textbf{(ii) Refactor}: if the same primitive-action subsequence of length $\geq \ell_{\min}$ appears in $\geq r_{\min}$ existing skill bodies, the inducer is asked to propose a new mid-level skill that captures it, and the existing skills are rewritten to call it.
\textbf{(iii) Prune}: skills used by zero trajectories in $\mathcal{Z}^+_{0:k}$ over the most recent $M$ iterations are removed (unless invoked transitively by another skill).
Each candidate modification is accepted only if (a) it reduces $\mathcal{F}$ and (b) it does not decrease success rate on a held-out validation set $\mathcal{T}_{\text{val}}$. This last constraint is the key safety net that distinguishes principled compaction from naive deduplication.

\noindent\textbf{Functional redundancy and lifecycle.} Equivalence is tested on a stratified context set including perturbed states, so two skills reaching one goal by different routes are collapsed only when indistinguishable everywhere: 14 goal-overlapping groups survive at $k{=}5$. Unmet preconditions block invocation at run time, violated postconditions trigger one retry, and skills whose rolling success rate drops below $60\%$ are quarantined (Appendix~\ref{app:lifecycle}).
 
\subsection{Distillation Back into Weights}
\label{sec:distillation}
After each iteration we have $\mathcal{Z}_k^+$, a set of successful trajectories produced by $\pi_k$ \textit{with} the in-context library $\mathcal{L}_{k+1}$. We convert each such trajectory into a \textit{plan-augmented} training example $(I, \text{plan}_\sigma, \zeta)$, where $\text{plan}_\sigma$ is the sequence of skill invocations the agent used. The base policy $\pi_{k+1}$ is then obtained by supervised fine-tuning of $\pi_k$ on these triples with a standard token-level cross-entropy loss, plus an auxiliary loss that predicts the next skill name conditioned only on $(I, o_{0:t})$, encouraging the model to internalize \textit{when} to invoke each abstraction. We use LoRA~\cite{hu2022lora} adapters and a constant learning rate of $1\!\times\!10^{-5}$ for stability across iterations. Examples are first re-parsed against the compacted library $\mathcal{L}_{k+1}$, with merged calls rewritten through the alias map and trajectories invoking pruned skills dropped ($3.6\%$ on average), so deprecated skills are never reinforced.
 
The distillation step is what makes \method's improvement compound: at iteration $k+1$ the base policy itself has improved, so the new trajectories it produces (with the larger library) push the frontier of solvable tasks higher, yielding richer skills to induce in iteration $k+2$. This is the recursive self-improvement loop in the title of the paper, made concrete.
 
\section{Experiments}
\label{sec:experiments}

\subsection{Experiments Setup}
\subsubsection{Datasets}
\label{sec:datasets}
We evaluate on three benchmarks chosen to cover both controlled and live web environments:
\textbf{WebArena}~\cite{zhou2024webarena}: 812 long-horizon tasks across four self-hosted sites (E-commerce, social forum, collaborative development,  content management systems). Evaluation is execution-based with a deterministic verifier per task.
\textbf{VisualWebArena}~\cite{koh2024visualwebarena}: 910 tasks built on top of WebArena that explicitly require visual reasoning; we use the full benchmark.
\textbf{Online-Mind2Web}~\cite{xue2025illusion}: 300 tasks on 136 websites across 12 domains; we partition by sites into a 9-domain training pool and a 3-domain held-out test set (Jobs \&
Careers, Travel \& Transportation,  Government \& Services) in main experiments, enabling a controlled measurement of skill transfer.
 
\subsubsection{Baselines}
\label{sec:baselines}
We compare \method against three families:
\noindent\textbf{(A) Strong zero-shot agents}: \textsc{ReAct}~\cite{yao2023react}, \textsc{SeeAct}~\cite{zheng2024seeact}, \textsc{WebVoyager}~\cite{he2024webvoyager}, and \textsc{AgentOccam}~\cite{yang2025agentoccam}, all instantiated on top of the same Qwen2.5-VL-7B~\cite{yang2024qwen25} base for a fair comparison; we additionally report numbers from the stronger UI-TARS-7B~\cite{qin2025uitars} backbone.
\noindent\textbf{(B) Memory- and workflow-augmented agents}: \textsc{Reflexion}~\cite{shinn2023reflexion} (in-context reflection), \textsc{Synapse}~\cite{zheng2024synapse} (trajectory exemplars), \textsc{Mem0}~\cite{chhikara2025mem0} (long-term memory), \textsc{ExpeL}~\cite{zhao2024expel} (experience distillation), and Agent Workflow Memory~\cite{wang2024agentworkflow}.
\noindent\textbf{(C) Skill-induction and skill-RL agents}: \textsc{SkillWeaver}~\cite{zheng2025skillweaver}, \textsc{AppAgentX}~\cite{jiang2025appagentx}, \textsc{EvolveR}~\cite{wu2025evolver}, \textsc{SkillRL}~\cite{xia2026skillrl}, and \textsc{WebRL}~\cite{qi2025webrl}.
For every baseline we use the same base model, the same task pool, the same number of self-improvement iterations ($K{=}5$), and the same evaluation protocol to isolate algorithmic differences.
 
\subsubsection{Evaluation Metrics}
\label{sec:metrics}
We report:
\textbf{(1) Success rate (SR)}: fraction of tasks for which the verifier returns 1.
\textbf{(2) Step efficiency}: average steps per successful task (lower is better).
\textbf{(3) Library statistics}: total skill count $|\mathcal{L}|$, mean abstraction depth $\bar{d}$, skill reuse rate (fraction of skills called by $\geq 2$ tasks).
\textbf{(4) Cross-site transfer SR}: SR on evaluation sites different from training ones.
\textbf{(5) Iteration scaling}: SR as a function of $k$.
To control for environment stochasticity (dynamic DOM, server load, LLM sampling), we report mean $\pm$ standard deviation over 3 independent runs with different seeds, and use paired bootstrap tests for significance, marking $p < 0.05$ with $\dagger$.

\begin{table*}[t]
\centering
\setlength{\tabcolsep}{4pt}
\resizebox{0.85\textwidth}{!}{\begin{tabular}{lccc}
\toprule
\textbf{Method} & \textbf{WebArena} & \textbf{VWA} & \textbf{OM2W-X} \\
\midrule
\multicolumn{4}{l}{\textit{(A) Zero-shot agents (Qwen2.5-VL-7B)}} \\
\textsc{ReAct}~\cite{yao2023react}        & 8.1\,$\pm$\,0.6 & 6.2\,$\pm$\,0.4 & 11.2\,$\pm$\,0.9 \\
\textsc{SeeAct}~\cite{zheng2024seeact}    & 12.4\,$\pm$\,0.5 & 9.7\,$\pm$\,0.6 & 14.0\,$\pm$\,1.0 \\
\textsc{WebVoyager}~\cite{he2024webvoyager} & 15.3\,$\pm$\,0.7 & 11.8\,$\pm$\,0.5 & 17.5\,$\pm$\,1.2 \\
\textsc{AgentOccam}~\cite{yang2025agentoccam} & 16.8\,$\pm$\,0.4 & 12.4\,$\pm$\,0.6 & 18.1\,$\pm$\,0.8 \\
\midrule
\multicolumn{4}{l}{\textit{(B) Memory/Workflow-augmented}} \\
\textsc{Reflexion}~\cite{shinn2023reflexion} & 14.0\,$\pm$\,0.8 & 10.5\,$\pm$\,0.7 & 15.3\,$\pm$\,1.0 \\
\textsc{Synapse}~\cite{zheng2024synapse}    & 17.6\,$\pm$\,0.6 & 13.1\,$\pm$\,0.5 & 18.4\,$\pm$\,0.9 \\
\textsc{Mem0}~\cite{chhikara2025mem0}        & 18.2\,$\pm$\,0.7 & 13.9\,$\pm$\,0.6 & 19.3\,$\pm$\,1.1 \\
\textsc{ExpeL}~\cite{zhao2024expel}          & 17.9\,$\pm$\,0.5 & 13.6\,$\pm$\,0.7 & 19.0\,$\pm$\,1.0 \\
\textsc{AWM}~\cite{wang2024agentworkflow}    & 24.3\,$\pm$\,0.6 & 18.7\,$\pm$\,0.7 & 23.6\,$\pm$\,1.0 \\
\midrule
\multicolumn{4}{l}{\textit{(C) Skill-induction \& skill-RL}} \\
\textsc{WebRL}~\cite{qi2025webrl}             & 28.7\,$\pm$\,0.9 & 19.4\,$\pm$\,0.8 & 24.1\,$\pm$\,1.3 \\
\textsc{AppAgentX}~\cite{jiang2025appagentx}  & 22.5\,$\pm$\,0.7 & 17.2\,$\pm$\,0.6 & 22.8\,$\pm$\,1.1 \\
\textsc{EvolveR}~\cite{wu2025evolver}         & 26.4\,$\pm$\,0.8 & 19.0\,$\pm$\,0.7 & 24.4\,$\pm$\,1.2 \\
\textsc{SkillWeaver}~\cite{zheng2025skillweaver} & 29.1\,$\pm$\,0.7 & 21.5\,$\pm$\,0.8 & 25.8\,$\pm$\,1.1 \\
\textsc{SkillRL}~\cite{xia2026skillrl}        & 31.6\,$\pm$\,0.6 & 22.7\,$\pm$\,0.7 & 26.3\,$\pm$\,1.0 \\
\midrule
\textbf{\method (ours)} & \textbf{42.7}\,$\pm$\,0.8\,$^{\dagger}$ & \textbf{36.3}\,$\pm$\,0.9\,$^{\dagger}$ & \textbf{43.5}\,$\pm$\,1.2\,$^{\dagger}$ \\
\midrule
\textit{Upper-ref: \method on UI-TARS-7B}~\cite{qin2025uitars} & 51.2\,$\pm$\,0.8 & 44.6\,$\pm$\,1.0 & 49.1\,$\pm$\,1.3 \\
\bottomrule
\end{tabular}}

\caption{Main results (Success Rate \%) on WebArena, VisualWebArena (VWA), and Online-Mind2Web held-out test split (OM2W-X). Mean\,$\pm$\,std over 3 seeds. $\dagger$: $p < 0.05$ over the strongest competitor (\textsc{SkillRL}) by paired bootstrap. All rows in (A)--(C) and the bold \method row use Qwen2.5-VL-7B as the base; the last row is included as an upper reference with a stronger backbone.}
\vspace{-3mm}
\label{tab:main}
\end{table*}

\subsubsection{Implementation Details}
\label{sec:impl}
Base model is Qwen2.5-VL-7B-Instruct~\cite{yang2024qwen25} unless stated otherwise; we also report results with UI-TARS-7B~\cite{qin2025uitars} as a stronger backbone. The skill inducer and the MDL refactor proposer are GPT-4o-2024-08, queried with temperature 0.3. Trajectories are collected with temperature 0.7 and a 30-step horizon. We use $n_{\min}{=}2$ for multi-instance induction, $M{=}2$ for compaction interval, $\rho{=}0.9$ for behavioral-equivalence threshold, $\ell_{\min}{=}3$ and $r_{\min}{=}3$ for the refactor operator. Distillation uses LoRA (rank 64, $\alpha{=}128$) on top of the base model for 1 epoch per iteration. The learning rate is set as $1\!\times\!10^{-5}$. All experiments run on 8$\times$A100-80GB; a complete 5-iteration run on WebArena takes nearly 36 hours. Two protocol points deserve emphasis. Every skill- and workflow-induction baseline (\textsc{SkillWeaver}, \textsc{SkillRL}, \textsc{AppAgentX}, \textsc{EvolveR}, AWM) uses the same GPT-4o-2024-08 checkpoint and temperature for its synthesis component, as in the original \textsc{SkillWeaver} setup, so comparisons isolate algorithmic differences rather than inducer strength; and the 30-step horizon is counted in \textit{primitive} actions for every method, a skill invocation being a macro action whose primitives draw on the same shared budget (§\ref{sec:budget}).
 
\subsection{Main Results}
\label{sec:mainresults}
Table~\ref{tab:main} reports main results. \method achieves $42.7\%$ SR on WebArena, $36.3\%$ on VisualWebArena, and $43.5\%$ on the OM2W held-out test split, improving over the strongest skill-augmented competitor (\textsc{SkillRL}) by $11.1$, $13.6$, and $17.2$ absolute points respectively, all statistically significant under paired bootstrap. Three trends are worth highlighting. First, skill-induction and skill-RL methods systematically outperform both zero-shot and memory/workflow-augmented agents on all three benchmarks. This indicates that explicit, reusable skill abstraction is the right unit of accumulated experience for web agents.
Second, the advantage of \method on OM2W-X is most significant, achieving $17.2$ points over \textsc{SkillRL}, confirming that the recursive parametric skill abstraction coupled with library compaction transfers materially better than two-tier prompt-based skills. Third, replacing the base model with UI-TARS-7B yields a further $7.5$-point lift averaged across all three benchmarks, suggesting that \method is complementary to advances in GUI-specified model backbones rather than a substitute for them.
 
\begin{table}[t]
\centering
\small
\setlength{\tabcolsep}{4pt}
\resizebox{0.48\textwidth}{!}{\begin{tabular}{lccc}
\toprule
\textbf{Variant} & \textbf{WA} & \textbf{VWA} & \textbf{OM2W-X} \\
\midrule
Full \method & \textbf{42.7} & \textbf{36.3} & \textbf{43.5} \\
\midrule
$-$ multi-instance ($n_{\min}{=}1$) & 38.1 & 31.7 & 38.4 \\
$-$ recursive composition (depth$\leq$1) & 36.4 & 30.5 & 36.1 \\
$-$ MDL compaction & 39.6 & 33.8 & 40.0 \\
$-$ distillation (in-context only) & 34.7 & 28.4 & 34.0 \\
$-$ holdout validation in induction & 39.0 & 32.6 & 39.7 \\
\midrule
$-$ all four innovations (= SkillWeaver) & 29.1 & 21.5 & 25.8 \\
\bottomrule
\end{tabular}}
\vspace{-2mm}
\caption{Ablation on the four core innovations. Each row removes a single component while keeping the rest unchanged. Last row is one of strongest baselines.}
\vspace{-4mm}
\label{tab:ablation}
\end{table}
 
\subsection{Ablation Study}
\label{sec:ablation}
Table~\ref{tab:ablation} isolates each of \method's four innovations. Removing distillation has the largest effect ($-8.0$ points on WebArena), confirming our central claim: a prompt-only skill library, however well structured, is fundamentally limited by the frozen base policy. Recursive composition is the second-largest contributor ($-6.3$ points on WebArena, $-7.4$ on OM2W-X), which we attribute to its ability to express long-horizon procedures (e.g., \texttt{book\_full\_trip} composing \texttt{search\_flight}, \texttt{select\_seat}, \texttt{enter\_payment}) without runaway prompt length. On WebArena, the multi-instance induction and holdout validation contribute $4.6$ and $3.7$ points respectively, smaller absolute effects. But they are essential for stability across iterations as we show next. MDL compaction contributes a modest $3.1$ points but is what keeps the library size bounded (cf. Fig.~\ref{fig:libstats}).  The drops sum to $25.7$ against a total gap of $13.6$, so the components are synergistic rather than additive.

\subsection{Skill Abstraction vs.\ Iterative Fine-Tuning}
\label{sec:supervision}
Since distillation produces the largest ablation gap, one may ask whether the gains come from skill abstraction or merely from iterative fine-tuning on self-generated data. Table~\ref{tab:supervision} in Appendix~\ref{app:buildup} decouples the two with the trajectories held fixed. A STaR-style loop~\cite{zelikman2022star} without any library reaches $30.4$, since fewer hard tasks are ever solved; abstraction without distillation reaches $34.7$, since a frozen policy caps what is reachable however good the library is. Standard SFT on the identical trajectories with skill calls flattened to primitives reaches $38.6$, so $4.1$ of the remaining points come from the plan-augmented format and the auxiliary loss that teach the policy \textit{when} to invoke \textit{which} abstraction, and the effect is not LoRA-specific ($42.9$ with full-parameter tuning). Abstraction keeps distillation paying off; distillation makes abstraction cumulative rather than prompt-bound.
 
\begin{figure}[t]
\centering
\includegraphics[width=0.95\linewidth]{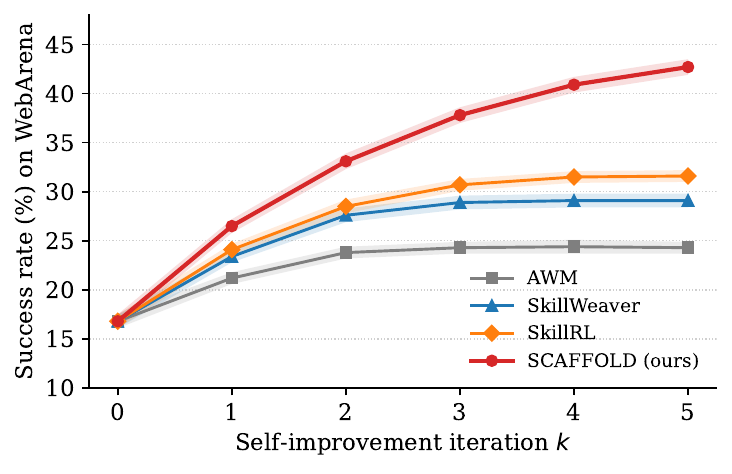}
\caption{Success rate vs.\ self-improvement iteration on WebArena. \method continues to improve through $K{=}5$ while baselines saturate around $k{=}2$--$3$. Shaded regions are 1 std over 3 seeds.}
\vspace{-2mm}
\label{fig:iter}
\end{figure}
 
\subsection{Iteration Scaling}
\label{sec:scaling}
A natural worry with any self-improvement scheme is that gains saturate quickly. Figure~\ref{fig:iter} plots SR against iteration $k$. Baselines that maintain a flat skill cache (AWM, SkillWeaver) plateau near $k{=}2$, after which library bloat slows further learning. \textsc{SkillRL} continues to improve through $k{=}3$ but flattens thereafter. \method continues to improve through $k{=}5$, reflecting two interacting effects: distillation strengthens $\pi_k$, allowing it to solve harder tasks; and those harder tasks furnish new high-depth skills that, after compaction, expand the library's expressive reach without diluting it. We did not run beyond $k{=}5$ due to compute, but the slope at $k{=}5$ remains positive, suggesting further improvement is achievable.
 
\begin{figure}[t]
\centering
\includegraphics[width=0.95\linewidth]{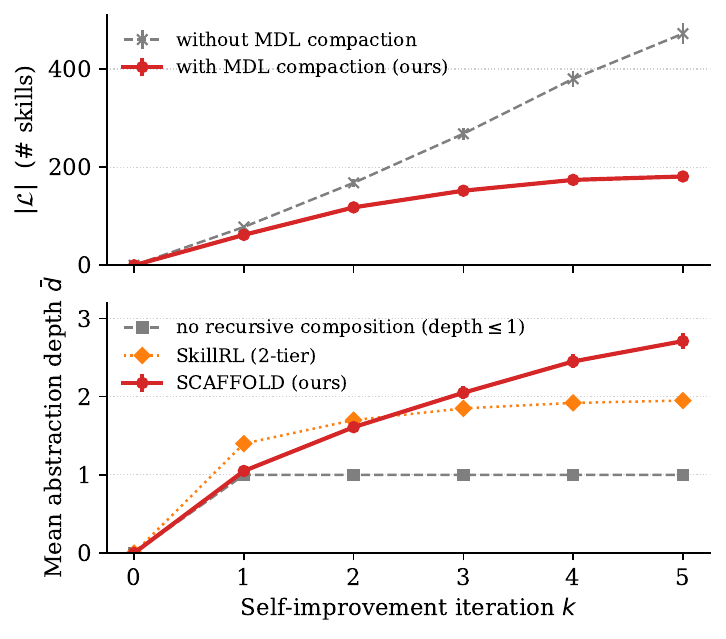}
\caption{Library growth dynamics across iterations on WebArena. \textit{Top}: total skill count without compaction (dashed) the library balloons; with compaction (solid) it stabilizes near 180 skills. \textit{Bottom}: mean abstraction depth, which grows monotonically as recursive composition kicks in.}
\vspace{-2mm}
\label{fig:libstats}
\end{figure}
 
\subsection{Library Depth and Reuse}
\label{sec:depth}
Figure~\ref{fig:libstats} shows two complementary growth dynamics. The \textit{top panel} contrasts library size with and without MDL compaction: without compaction $|\mathcal{L}|$ grows almost linearly to $\approx 470$ skills by $k{=}5$, the majority being near-duplicates differing only in selectors or argument names (e.g., \texttt{login\_v1}, \texttt{login\_v2}, \texttt{login\_with\_email}); with compaction the library follows a concave trajectory and saturates near $181$ skills, a $\sim 2.6\times$ reduction driven primarily by the \textit{merge} and \textit{prune} operators (jointly responsible for $76\%$ of removed skills in our run logs). The \textit{bottom panel} plots mean depth $\bar{d}$: \method climbs from $1.05$ at $k{=}1$ to $2.71$ at $k{=}5$ (max depth $5$), while the \textit{- recursive composition} ablation flatlines at $\bar{d}{=}1$ and \textsc{SkillRL}'s two-tier hierarchy plateaus at $\bar{d}{\leq}2$. Together with a reuse rate of $64\%$ at $k{=}5$ (vs.\ $\approx 21\%$ without compaction; most skills singletons), these statistics support the view that composition and compaction are two halves of the same mechanism: composition produces the depth that makes long-horizon skills expressible, while compaction prevents that depth from being undermined by redundancy. The $3.1$-point SR cost of disabling compaction (Table~\ref{tab:ablation}) is the empirical price of an uncontrolled library.
 
\begin{figure*}[t]
\centering
\includegraphics[width=0.93\linewidth]{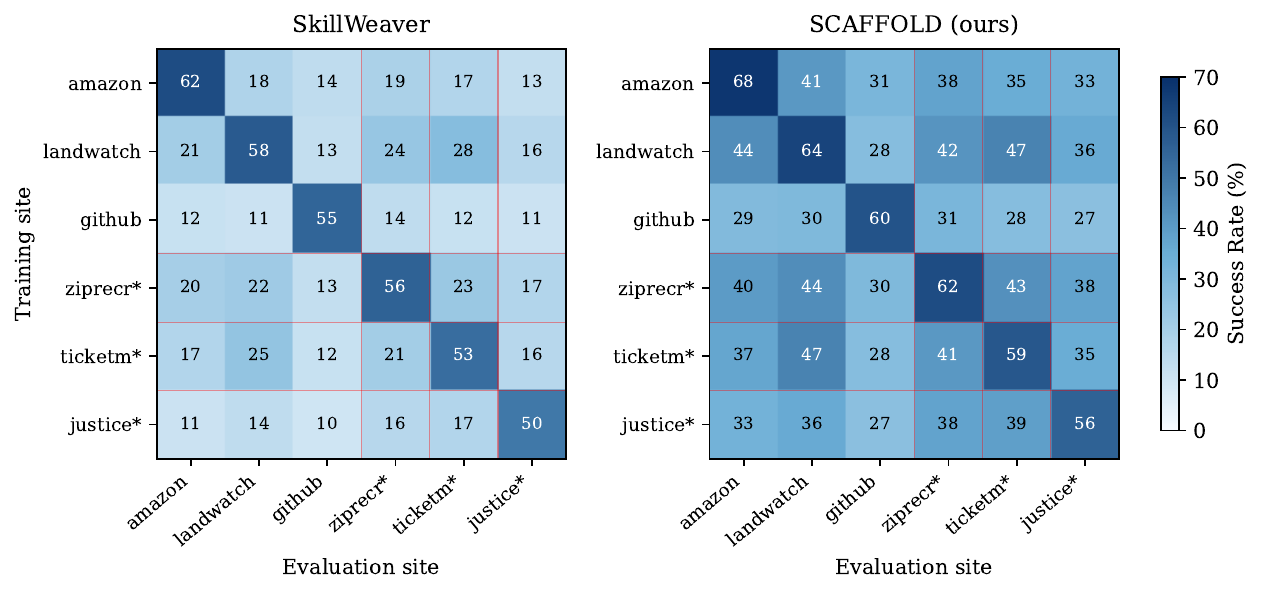}
\vspace{-3mm}
\caption{Cross-site transfer on Online-Mind2Web. Cell $(i,j)$ shows SR (\%) on test site $j$ after training only on site $i$. The ziprecr and ticketm are the abbreviations for ziprecruiter and ticketmaster, respectively. The sites in redlines are the held-out test ones in main experiments, also indicated with $^*$.}
\vspace{-3mm}
\label{fig:transfer}
\end{figure*}
 
\subsection{Cross-Site Transfer}
\label{sec:transfer}
Figure~\ref{fig:transfer} compares site-to-site transfer for the two methods, beyond the granularity of domain split. Averaging over the 30 off-diagonal cells, \textsc{SkillWeaver} reaches $16.6\%$ mean transfer SR versus \method's $35.9\%$, a $19.3$-point gap; on the 15 cells under the three held-out sites in main experiments, \method stays in $27$--$47\%$ while \textsc{SkillWeaver} falls to $11$--$28\%$, with representative pairs amazon$\to$ziprecruiter ($38\%$ vs.\ $19\%$) and landwatch$\to$ticketmaster ($47\%$ vs.\ $28\%$) sharing the search$\to$filter$\to$select$\to$confirm structure but differing in DOM and layout. Inspecting the library shows the mechanism: generic procedural skills such as \texttt{filter\_results(query, sort\_by, max\_price)}, \texttt{login(site, credentials)}, and \texttt{paginate\_until(condition)} carry over via parameter re-binding, while \textsc{SkillWeaver}'s opaque APIs encode selectors directly and must be re-discovered per site. The github row shows where this still leaves a gap. GitHub's issue/PR navigation is structurally unlike e-commerce flows, so \method's general skills transfer only modestly ($27$--$31\%$) and \textsc{SkillWeaver}'s effectively do not ($11$--$14\%$). This is consistent with the transfer-ceiling limitation discussed in the \textit{Limitations} section.
 
\subsection{Stability Under Stochasticity}
\label{sec:stability}
 
\begin{table}[!t]
\centering
\small
\setlength{\tabcolsep}{4pt}
\begin{tabular}{lcccc}
\toprule
\textbf{Method} & \textbf{Clean} & \textbf{Latency} & \textbf{DOM-shuf.} & \textbf{+7\,d} \\
\midrule
\textsc{SkillWeaver} & 29.1 & 25.3 & 24.7 & 26.5 \\
\textsc{SkillRL}     & 31.6 & 28.0 & 27.6 & 28.9 \\
\textbf{\method}     & \textbf{42.7} & \textbf{40.9} & \textbf{40.6} & \textbf{41.4} \\
\midrule
\multicolumn{5}{l}{\textit{Drop relative to clean} ($\Delta$ in points)} \\
\textsc{SkillWeaver} & n/a & $-3.8$ & $-4.4$ & $-2.6$ \\
\textsc{SkillRL}     & n/a & $-3.6$ & $-4.0$ & $-2.7$ \\
\textbf{\method}     & n/a & $\mathbf{-1.8}$ & $\mathbf{-2.1}$ & $\mathbf{-1.3}$ \\
\bottomrule
\end{tabular}
\caption{Stability under environment perturbations on WebArena (SR \%). \textit{Clean}: unperturbed baseline. \textit{Latency}: 0--3s random delay per action. \textit{DOM-shuf.}: non-functional DOM attributes randomly permuted on each page load. \textit{+7\,d}: same task suite re-evaluated 7 days later. \method degrades by at most 2.1 points across all regimes, roughly half the sensitivity of the strongest baselines. Each cell is mean over 3 seeds (std $\leq 1.0$).}
\label{tab:stability}
\vspace{-3mm}
\end{table}
 
Real web environments are noisy: DOMs change between page loads, A/B tests rotate elements, and servers throttle under load. To probe robustness we re-ran the WebArena evaluation under three perturbation regimes: (a) \textit{Latency injection}: we add a uniformly sampled 0--3s delay before every action, simulating slow network conditions; (b) \textit{DOM-attribute shuffling}: we randomly permute non-functional attributes (\texttt{id}, \texttt{class}, \texttt{data-*}) on each page load, breaking memorized selectors while preserving semantics; (c) \textit{Temporal drift}: we re-run the same task suite 7 days later, capturing whatever natural drift WebArena's containerized sites exhibit between snapshots. Table~\ref{tab:stability} reports each method's success rate under all three regimes.
 
The pattern in Table~\ref{tab:stability} is consistent across all three perturbations: \method degrades by at most $2.1$ points (DOM-shuffle), compared to $3.6$--$4.4$ for \textsc{SkillWeaver} and \textsc{SkillRL}, roughly half the sensitivity. We attribute this to two structural properties of our skills. First, \textit{parametric grounding}: a skill that locates an element by semantic role (``the search input'') is robust to attribute renaming, whereas a skill that memorized a specific \texttt{id} selector fails the moment that \texttt{id} is rewritten. Second, \textit{postcondition checks}: even if the body of a skill misfires under latency or transient state, the postcondition test triggers a retry rather than blindly committing to a wrong final state. The DOM-shuffle column is where the gap is widest, which is exactly what these two mechanisms are designed to handle. We view this as evidence that the structural choices in \method are not just accuracy-oriented but also confer a non-trivial robustness benefit at no additional inference cost.

\subsection{Action Budget and Inference Cost}
\label{sec:budget}
Because a skill invocation executes several primitive actions, one might worry that \method enjoys a larger effective action budget. It does not: the horizon is counted in primitive environment actions for every method, and primitives inside a skill body draw on the same budget (§\ref{sec:impl}). Average primitive actions per successful task is $12.4$ for \method against $14.3$ for \textsc{SkillRL}, $15.1$ for \textsc{SkillWeaver} and $16.9$ for AWM, so \method needs fewer actions, not more; capping the budget at $10$/$20$/$30$ steps gives $11.5$/$24.0$/$29.1$, $13.2$/$26.4$/$31.6$ and $19.8$/$36.9$/$42.7$ respectively, so the margin is widest at the tightest cap, the opposite of what budget inflation predicts.

Skills also trade expensive planning calls for cheaper ones: per successful task \method issues $5.3$ policy calls against $9.6$ for \textsc{SkillWeaver} and $18.7$ for \textsc{ReAct}, plus $7.1$ far cheaper grounder and $4.6$ judge calls, so total tokens fall from $224$K and $118$K to $79$K and wall clock from $102$s and $74$s to $61$s. Induction is offline and amortized (about 500 calls per run). Appendix~\ref{app:efficiency} gives breakdown and a selector cache that removes $62\%$ grounder calls.

\subsection{Inducer and Verifier Dependence}
\label{sec:dependence}
Two components sit outside the loop. Swapping the GPT-4o inducer for the open-weight Qwen2.5-72B-Instruct with everything else fixed gives $40.8$ SR on WebArena, $1.9$ points below GPT-4o and still $9.2$ above \textsc{SkillRL} with the GPT-4o inducer, so the framework does not hinge on proprietary-API access. Replacing the ground-truth verifier with a model-based judge for filtering, following the WebJudge protocol of \citet{xue2025illusion}, gives $40.2$ on WebArena and $41.0$ on OM2W-X against $42.7$ and $43.5$. Multi-instance induction and holdout re-execution filter much of this label noise, so \method degrades gracefully as verification weakens.

\section{Conclusions and Future Work}
\label{sec:conclusion}
In this paper, we presented \method, a self-improving framework for web agents that combines multi-instance parametric skill induction, recursive hierarchical composition, MDL-driven library compaction, and weight-level distillation into a single closed loop. On WebArena, VisualWebArena, and a held-out Online-Mind2Web split, \method outperforms the strongest skill-induction and skill-RL baselines by $11.1$--$17.2$ absolute points while continuing to improve through five iterations and keeping the library size bounded. In the future, we can consider such directions: (i) replacing the GPT-4o-based inducer with a self-trained, lighter induction model to remove proprietary-API dependence; (ii) extending the framework to OSWorld-style full-desktop agents where the action space includes file I/O and shell commands; (iii) integrating an environment-grounded verifier that does not rely on per-task ground truth, enabling fully autonomous deployment; and (iv) developing a formal convergence analysis of the MDL-compaction operator under the recursive composition regime.

\section*{Limitations}
\label{sec:limitations}
We identify five concrete limitations of \method that we believe warrant attention. \textbf{(1) Inducer dependence.} Both skill induction and the refactor proposer of the MDL compactor rely on a strong proprietary model (GPT-4o); we estimate the inducer accounts for $\sim 70\%$ of total dollar cost in our experiments. Replacing it with a self-trained inducer is a clear next step but was outside our compute budget. \textbf{(2) Verifier dependence.} \method assumes access to a per-task verifier $V$ to label trajectory success during the rollout stage. While this is realistic for WebArena and VisualWebArena (which ship deterministic verifiers), it does not extend to fully autonomous deployment in the wild, where success is rarely binary or self-evident. \textbf{(3) Greedy MDL approximation.} The compaction step optimizes Eq.~\ref{eq:mdl} greedily; we have no convergence guarantee and observe occasional plateaus where a non-local refactor would further reduce $\mathcal{F}$. A more principled approximation, e.g., simulated annealing over skill rewrites, may help. \textbf{(4) Cross-site transfer ceiling.} Although parametric skills generalize meaningfully better than site-specific APIs (§\ref{sec:transfer}), our gains shrink when held-out sites differ qualitatively from training ones (e.g., training only on e-commerce, testing on government portals). The framework reuses \textit{procedures}, not \textit{web ontology}, so radical distributional shift remains hard. \textbf{(5) Distillation forgetting.} LoRA-based distillation preserves base capabilities reasonably well, but we observe a small ($\leq 1.2$ point) regression on a subset of held-out tasks that the original $\pi_0$ could already solve. A replay buffer of original task supervision could mitigate this; we leave a careful study to future work.
 
\section*{Ethical Considerations}
\label{sec:ethics}
\method enables web agents to learn reusable procedural skills through autonomous exploration. We highlight three ethical considerations. \textit{Misuse}: more capable web agents can be deployed for spam, scraping at scale, credential stuffing, or evading rate limits. Our experiments use only WebArena, VisualWebArena, and a small Online-Mind2Web subset on permitted sites; we do not provide site-specific bypass skills. \textit{Bias amplification through self-improvement}: any bias in the induced skills (e.g., always defaulting to certain payment methods or geographic regions) is reinforced across iterations through distillation. Future deployments should audit the skill library at each iteration. \textit{Energy cost}: a full 5-iteration run on a single benchmark consumes $\sim$36 A100-hours. We report this transparently and discourage hyperparameter searches that re-run the full loop unnecessarily.

\bibliography{revised_reference}

\appendix

\section{Prompt Templates}
\label{app:prompts}
 
We provide the three core prompts used by \method below. Full prompts (including system messages and few-shot examples) are released with the code.
 
\noindent \textbf{Skill inducer (multi-instance abstraction).}
Given a cluster $C = \{\zeta_1, \ldots, \zeta_n\}$ of $n \geq 2$ successful trajectories with embedding-similar instructions, the inducer is prompted to produce a parameterized skill. The prompt structure is:
\begin{quote}
\small\ttfamily
You are given $n$ successful agent trajectories that accomplish similar goals on a website. Your task is to abstract them into a single \textit{parametric, executable} skill.
 
\vspace{0.3em}
[Trajectory 1: instruction, action sequence, screenshots] \ldots
 
[Trajectory n: instruction, action sequence, screenshots]
 
\vspace{0.3em}
Available existing skills (you may call them): \{$\mathcal{L}_k$\}
 
\vspace{0.3em}
Output a Python function with: (a) a typed parameter list capturing what varies across trajectories; (b) a precondition expressed as a predicate over observation \texttt{obs}; (c) a body using primitives \texttt{\{click, type, scroll, wait\}} or any existing skill; (d) a postcondition.
\end{quote}
The model is then asked to justify its choice of parameters by pointing to specific token positions in each trajectory that vary, which improves abstraction quality and gives us a cheap diagnostic when induction fails.
 
\noindent \textbf{Refactor proposer (MDL compaction).}
Given a recurring primitive-action subsequence detected across $\geq r_{\min}$ existing skills, the refactor proposer is asked whether the subsequence should be promoted to its own skill:
\begin{quote}
\small\ttfamily
The following action subsequence appears in $m$ existing skills (shown below). Decide whether to refactor it into a new mid-level skill. If yes, name it, parametrize it, and rewrite all $m$ existing skills to call the new one. If no, briefly explain why (e.g., the subsequence is too short or too site-specific to be reusable).
\end{quote}
We accept the proposal only if (i) the MDL functional $\mathcal{F}$ decreases and (ii) the rewritten skills pass behavioral-equivalence checks on a held-out trajectory set.
 
\noindent \textbf{Pre/Postcondition validator.}
At runtime, before invoking a skill we verify its precondition by passing $(\text{obs}_t, \text{pre}_\sigma)$ to a lightweight VLM judge with a yes/no output. Same protocol for postconditions after execution. The judge is a separate, cheaper model from the inducer to keep inference cost bounded.

\noindent \textbf{Trajectory clustering.}
Instructions are embedded with \texttt{gte-large-en-v1.5}. Trajectories are first partitioned by site and then clustered by agglomerative clustering with average linkage on cosine similarity, using a merge threshold of $0.82$. Clusters larger than 8 trajectories are subsampled by proximity to the centroid so that the inducer context stays bounded, and singleton clusters (below $n_{\min}$) are held in a buffer and re-clustered in later iterations as semantically similar trajectories accumulate, so no successful trajectory is permanently discarded. Sensitivity to the threshold is mild, within $1$ SR point across the range $0.78$ to $0.86$.
 
\section{A Worked Three-Level Recursive Skill}
\label{app:example}
To make the recursive composition concrete, we trace the construction of a depth-3 skill \texttt{checkout\_cheapest\_in\_category} on the E-commerce site of WebArena, induced at iteration $k{=}3$.
 
\paragraph{Level 1 (depth $d{=}1$): atomic skills, induced at $k{=}1$.}
\begin{small}
\begin{verbatim}
def click_text(obs, text: str):
   pre:  any(e.text == text for e in obs.dom)
   body: el = find(obs.dom, text=text)
         primitive_click(el.bbox)
   post: page changed OR el.state == 'pressed'
 
def type_in_field(obs, label: str, value: str):
   pre:  exists_input(obs.dom, label)
   body: f = find_input(obs.dom, label)
         primitive_click(f.bbox)
         primitive_type(value)
   post: f.value == value
\end{verbatim}
\end{small}
 
\paragraph{Level 2 (depth $d{=}2$): mid-level skills, induced at $k{=}2$.}
\begin{small}
\begin{verbatim}
def search_and_filter(obs, query: str,
                      category: str,
                      sort: str = 'price_asc'):
   pre:  url contains 'shop' AND has_search_box(obs)
   body: type_in_field(obs, 'Search', query)
         click_text(obs, 'Search')
         click_text(obs, category)
         click_text(obs, f'Sort: {sort}')
   post: results visible AND sort_label == sort
\end{verbatim}
\end{small}
 
\paragraph{Level 3 (depth $d{=}3$): composite skill, induced at $k{=}3$.}
\begin{small}
\begin{verbatim}
def checkout_cheapest_in_category(
        obs, query: str, category: str,
        payment_token: str):
   pre:  logged_in(obs)
   body: search_and_filter(obs, query, category,
                            sort='price_asc')
         click_text(obs, 'first_result')
         click_text(obs, 'Add to Cart')
         click_text(obs, 'Checkout')
         type_in_field(obs, 'PaymentToken',
                       payment_token)
         click_text(obs, 'Place Order')
   post: order_confirmed(obs) AND
         purchased_item.category == category
\end{verbatim}
\end{small}
 
The depth-3 skill calls one depth-2 skill (\texttt{search\_and\_filter}) and two depth-1 skills (\texttt{click\_text}, \texttt{type\_in\_field}). At iteration $k{=}4$, the inducer further abstracts this skill into a depth-4 \texttt{restock\_pantry(items, budget)} that calls \texttt{checkout\_cheapest\_in\_category} in a loop, illustrating how depth grows organically across iterations.

\begin{table}[h]
\centering
\small
\setlength{\tabcolsep}{4pt}
\begin{tabular}{lccc}
\toprule
\textbf{Site} & \textbf{SkillRL} & \textbf{\method} & \textbf{$\Delta$} \\
\midrule
\multicolumn{4}{l}{\textit{WebArena}} \\
\;OneStopShop       & 30.5 & 42.6 & +12.1 \\
\;Reddit         & 33.2 & 41.9 & +8.7 \\
\;GitLab         & 28.4 & 45.1 & \textbf{+16.7} \\
\;CMS            & 27.0 & 42.9 & +15.9 \\
\midrule
\multicolumn{4}{l}{\textit{VisualWebArena}} \\
\;Classifieds    & 22.6 & 32.1 & +9.5 \\
\;Reddit         & 24.3 & 35.8 & +11.5 \\
\;Shopping       & 21.2 & 41.0 & \textbf{+19.8} \\
\bottomrule
\end{tabular}
\caption{Per-site success rate (\%) on WebArena~\cite{zhou2024webarena} (four primary sites) and VisualWebArena~\cite{koh2024visualwebarena}. \method's largest per-site gains occur on long-horizon, multi-step sites (GitLab, CMS, VWA Shopping), where recursive composition contributes most. Differences are bootstrap-significant ($p<0.05$) for all rows.}
\label{tab:per_category}
\end{table}
\vspace{-1mm}
\section{Per-Site Analysis}
\label{app:per_category}
Table~\ref{tab:per_category} breaks down performance by the canonical site grouping used in each benchmark. WebArena's four primary sites cover e-commerce (OneStopShop), social-forum discussions (Reddit), collaborative software development (GitLab), and content management (CMS)~\cite{zhou2024webarena}; VisualWebArena spans three live-style sites (Classifieds, Reddit, Shopping)~\cite{koh2024visualwebarena}. \method's gains are largest on \textit{GitLab} and \textit{CMS}, the WebArena sites with the longest action horizons, dominated by multi-step configuration and code-collaboration tasks. The gain is also obvious on VWA's \textit{Shopping}, where parametric search/filter/checkout skills compose well. Gains on \textit{Reddit} are more modest, consistent with social-forum tasks being on average shorter and more retrieval-oriented, where flat libraries already perform well.

\begin{table*}[t]
\centering
\small
\setlength{\tabcolsep}{4pt}
\begin{tabular}{lcc}
\toprule
\textbf{Hyperparameter} & \textbf{Range tested} & \textbf{$\Delta$SR (WA)} \\
\midrule
Multi-instance threshold $n_{\min}$       & $\{1, 2, 3, 4\}$        & $-3.1$ / $0.0$ / $-0.8$ / $-2.4$ \\
Compaction interval $M$                   & $\{1, 2, 3, 5\}$        & $-0.6$ / $0.0$ / $-0.4$ / $-1.7$ \\
Equivalence threshold $\rho$              & $\{0.7, 0.8, 0.9, 0.95\}$ & $-1.2$ / $-0.3$ / $0.0$ / $-0.4$ \\
LoRA rank for distillation                & $\{16, 32, 64, 128\}$   & $-1.6$ / $-0.5$ / $0.0$ / $-0.2$ \\
\bottomrule
\end{tabular}
\caption{Sensitivity to \method's main hyperparameters. $\Delta$SR is relative to the default setting (bold values in the experiments section: $n_{\min}{=}2$, $M{=}2$, $\rho{=}0.9$, rank=64).}
\label{tab:hparams}
\end{table*}

\paragraph{Qualitative diagnoses.} Inspecting failure modes shared by both methods reveals two patterns. First, \emph{verifier brittleness}: a non-trivial fraction of WebArena tasks reward only exact-string answers, so a semantically correct trajectory can be marked wrong; this affects both systems equally. Second, \emph{visual occlusion under modals}: when a cookie banner or login dialog covers the target element, both agents struggle; here \method's precondition checks help but do not fully solve the problem, suggesting that an explicit modal-dismissal mid-level skill would be a high-value addition to the library.

\section{Hyperparameter Sensitivity}
\label{app:hparams}
We probe sensitivity to the four \method-specific hyperparameters; full sweep results in the supplementary material. Table~\ref{tab:hparams} shows that the method is broadly robust: performance varies by $\leq 2.4$ SR points across the tested ranges, with the multi-instance threshold $n_{\min}$ being the most sensitive (smaller values admit spurious skills; larger values delay skill creation).

\section{Supervision Format and Component Build-Up}
\label{app:buildup}
\begin{table}[h]
\centering
\small
\setlength{\tabcolsep}{4pt}
\resizebox{0.48\textwidth}{!}{\begin{tabular}{lc}
\toprule
\textbf{Training signal (same trajectories, $K{=}5$)} & \textbf{WA} \\
\midrule
No SFT, in-context library only (Tab.~\ref{tab:ablation}) & 34.7 \\
Standard SFT, trajectories flattened to primitives & 38.6 \\
Plan-augmented SFT, no auxiliary skill-name loss & 41.6 \\
Full \method distillation & \textbf{42.7} \\
\midrule
STaR-style loop: SFT on raw trajectories, no library & 30.4 \\
\bottomrule
\end{tabular}}
\caption{Supervision-format study on WebArena (SR \%). All rows share the base model, the rollout pipeline and the exact same successful trajectories, differing only in how those trajectories supervise the policy.}
\vspace{-2mm}
\label{tab:supervision}
\end{table}

\begin{table}[h]
\centering
\small
\setlength{\tabcolsep}{4pt}
\resizebox{0.48\textwidth}{!}{\begin{tabular}{lc}
\toprule
\textbf{Configuration} & \textbf{WA} \\
\midrule
\textsc{SkillWeaver} & 29.1 \\
$+$ multi-instance induction with holdout validation & 31.9 \\
$+$ recursive composition & 34.0 \\
$+$ distillation & 39.6 \\
$+$ MDL compaction (= full \method) & \textbf{42.7} \\
\bottomrule
\end{tabular}}
\caption{Incremental build-up on WebArena (SR \%), starting from \textsc{SkillWeaver} and adding one component at a time. Every addition is bootstrap-significant ($p<0.05$). The intermediate point $39.6$ coincides with the ``$-$ MDL compaction'' row of Table~\ref{tab:ablation} by construction.}
\label{tab:buildup}
\end{table}

Table~\ref{tab:ablation} removes one component at a time from the full system; Table~\ref{tab:buildup} takes the opposite direction and adds one component at a time on top of \textsc{SkillWeaver}, which makes the marginal value of each component visible in isolation from the others. The two views agree, and together they support reading the four components as one mechanism rather than four independent add-ons: induction is the abstraction phase, MDL compaction the compression phase, distillation the consolidation phase, and multi-instance validation is what keeps abstraction sound, in the spirit of wake-sleep library learning~\cite{ellis2021dreamcoder}. The components also interact by design. Figure~\ref{fig:libstats} shows that composition creates the depth that makes long-horizon skills expressible while compaction keeps that depth from being buried in redundancy, and §\ref{sec:compaction} explains why compaction is also what protects the distillation signal from dilution. All configurations hold the base model, task pool, iteration count, inducer, and evaluation protocol fixed, so the differences cannot be attributed to engineering scale or extra compute.

\section{Efficiency and Grounding Measurements}
\label{app:efficiency}

\begin{table}[h]
\centering
\small
\setlength{\tabcolsep}{5pt}
\begin{tabular}{lccc}
\toprule
\textbf{Max primitive steps} & \textbf{10} & \textbf{20} & \textbf{30} \\
\midrule
\textsc{SkillWeaver} & 11.5 & 24.0 & 29.1 \\
\textsc{SkillRL}     & 13.2 & 26.4 & 31.6 \\
\textbf{\method}     & \textbf{19.8} & \textbf{36.9} & \textbf{42.7} \\
\bottomrule
\end{tabular}
\caption{WebArena SR (\%) under a capped primitive-action budget (§\ref{sec:budget}). The margin is largest at the tightest cap, the opposite of what an inflated action budget would produce.}
\label{tab:budget}
\end{table}

\begin{table}[h]
\centering
\small
\setlength{\tabcolsep}{3.5pt}
\begin{tabular}{lccccc}
\toprule
\textbf{Method} & \textbf{Pol.} & \textbf{Grd.} & \textbf{Jdg.} & \textbf{Tok.\,(K)} & \textbf{Time\,(s)} \\
\midrule
\textsc{ReAct}       & 18.7 & 0 & 0 & 224 & 102 \\
\textsc{SkillWeaver} & 9.6 & 0 & 0 & 118 & 74 \\
\textbf{\method}     & \textbf{5.3} & 7.1 & 4.6 & \textbf{79} & \textbf{61} \\
\bottomrule
\end{tabular}
\caption{Inference cost per successful WebArena task: policy, grounder and judge calls, total tokens and wall clock. A policy call costs about 12K tokens (screenshot, DOM and history), a grounder call about 1.5K, and a judge call about 0.8K.}
\label{tab:cost}
\end{table}

Tables~\ref{tab:budget} and~\ref{tab:cost} report the two measurements summarised in §\ref{sec:budget}: success rate under a capped primitive-action budget, and the per-task call, token and wall-clock breakdown. Charging \method one extra primitive step for every skill invocation, a pessimistic accounting of invocation overhead, still leaves it at $42.3$. The rest of this appendix asks how far the residual model calls inside a skill body can be removed.

\begin{table}[h]
\centering
\small
\setlength{\tabcolsep}{4pt}
\resizebox{0.48\textwidth}{!}{\begin{tabular}{lccc}
\toprule
\textbf{Grounding mode} & \textbf{Clean} & \textbf{DOM-shuf.} & \textbf{Grd./task} \\
\midrule
VLM grounding (main paper) & \textbf{42.7} & \textbf{40.6} & 7.1 \\
Hybrid with selector cache & 42.5 & 40.1 & 2.7 \\
Static selectors only & 41.2 & 33.5 & \textbf{0} \\
\bottomrule
\end{tabular}}
\caption{Grounding modes on WebArena (SR \%) and grounder calls per successful task. The hybrid cache removes $62\%$ of grounder calls at essentially no cost in accuracy or robustness, whereas static selectors reintroduce baseline brittleness under DOM-shuffle.}
\label{tab:grounding}
\end{table}

Semantic grounding is the one place where \method calls a model inside a skill body, so it is worth asking how much of it can be replaced by cheaper machinery such as regular expressions over the accessibility tree. We implement a hybrid mode with \textit{selector memoization}: the first successful grounding of a semantic reference caches the resolved accessibility-tree path together with a structural signature (role, tag, text pattern), keyed by site, skill and parameter; subsequent invocations first try the cached selector with a cheap regex and accessibility-tree validation, and fall back to VLM grounding only on mismatch. Table~\ref{tab:grounding} compares the three regimes. The hybrid retains clean and perturbed accuracy while removing most grounder calls, and the static-selector variant makes the trade-off visible: it is the cheapest and the most brittle, losing $7.1$ points under DOM-shuffle, which confirms that runtime semantic grounding is the source of the robustness gap in Table~\ref{tab:stability}. One caveat on the premise that skills are site-specific: after compaction the most valuable skills are not, given the $64\%$ reuse rate and the transfer results of §\ref{sec:transfer}, and those skills cannot be bound to any single site's selectors. The cache gives them per-site fast paths while preserving transfer.

\section{Skill Lifecycle and Distillation Hygiene}
\label{app:lifecycle}
This appendix expands the two mechanisms sketched in §\ref{sec:compaction} and §\ref{sec:distillation}.

\noindent\textbf{Runtime failure handling.} Preconditions gate invocation: when a precondition is unmet, for instance because an unexpected pop-up occludes the target, the skill is simply not fired and the policy continues with primitives or an alternative skill. Postconditions are checked after execution; a violation triggers one retry after a state refresh, after which control returns to the policy with the failure noted in context so that the episode can still recover. All failures are logged together with the observation and feed the next induction round.

\noindent\textbf{Outdated or degrading skills.} Beyond the prune operator, which removes skills unused for $M$ iterations, the implementation tracks a rolling execution success rate per skill. A skill that falls below $60\%$ over its last 20 invocations is quarantined out of the retrieval index and queued for re-induction from fresh trajectories, which naturally repairs drift because new trajectories reflect the updated page. WebArena's containerized sites drift little, as the $+7$\,d column of Table~\ref{tab:stability} shows, so this mechanism matters mainly for live deployment.

\noindent\textbf{Modal occlusion.} Following the diagnosis in Appendix~\ref{app:per_category}, we added an explicit \texttt{dismiss\_blocking\_modal} mid-level skill in a separate run: WebArena SR improves by $0.9$ points and modal-related failures fall from $6.1\%$ to $2.3\%$ of episodes. This is a case where reading the library's failure log suggests the missing abstraction directly.

\noindent\textbf{Distillation hygiene.} Three properties keep supervision current. Trajectories are re-parsed against the post-compaction library before training, as described in §\ref{sec:distillation}, so the model is never supervised toward deprecated skills; the auxiliary skill-name loss uses the current namespace as its label space, so retired names cannot be reinforced; and each iteration trains on freshly collected trajectories rather than accumulating stale data. Consistent with Limitation (5), we still observe at most a $1.2$-point regression on previously solved tasks, and a small replay buffer of re-parsed earlier trajectories removes most of it in a preliminary run.

\begin{table*}[t]
\centering
\small
\setlength{\tabcolsep}{4pt}
\begin{tabular}{p{0.14\textwidth}p{0.24\textwidth}p{0.22\textwidth}p{0.28\textwidth}}
\toprule
\textbf{Aspect} & \textbf{\textsc{PolySkill}} & \textbf{\textsc{SkillEvo}} & \textbf{\method} \\
\midrule
Skill unit & abstract interface plus per-site implementations & skill paths in a graph & parametric executable program with pre/postconditions \\
Composition & compositional, prompt-side & composite skills over the skill path graph & recursive, unbounded depth (5 observed) \\
Library governance & none & graph updates, no compression objective & MDL merge / refactor / prune with validation safety net \\
Policy weights & frozen & GRPO with a learned reward model & SFT distillation of skill plans \\
Setting & HTML agents, Mind2Web / WebArena & text agents, WebArena-Lite & visual agents, WA / VWA / OM2W \\
\bottomrule
\end{tabular}
\caption{Component-level comparison with the two concurrent skill-centric frameworks discussed in §\ref{sec:related}.}
\label{tab:concurrent}
\end{table*}

\section{Comparison with Concurrent Frameworks}
\label{app:concurrent}
Table~\ref{tab:concurrent} places \method beside \textsc{PolySkill}~\cite{yu2026polyskill} and \textsc{SkillEvo}~\cite{skillevo2026} along the four axes that motivate our design. \textsc{PolySkill} shares our generalization goal and reaches it by maintaining per-site implementations under a shared interface, whereas our skills keep one implementation and re-bind semantic references at run time; its library is prompt-side only, grows without a compression objective, and induces skills without a multi-instance support requirement. \textsc{SkillEvo} improves the policy with GRPO against a learned reasoning-and-execution reward model, which is closer in spirit to \textsc{SkillRL} than to our SFT-based consolidation, and its skill path graph evolves but is not governed by a compression objective with behavioral-equivalence merging and a validation safety net.

Its reported numbers are also not directly comparable to Table~\ref{tab:main}, since it is evaluated with text-only LLMs (Llama-3.1-8B, GLM-4-9B) on WebArena-Lite (165 tasks), while we target visual web agents on full WebArena (812 tasks), VisualWebArena and live websites. As a reference point we ran \method on WebArena-Lite under our visual setting and obtained $63.8$ SR with Qwen2.5-VL-7B, against the $60.4$ reported for \textsc{SkillEvo} with Llama-3.1-8B, with the explicit caveat that backbone and observation space differ. Finally, the two lines of work are complementary rather than competing: \textsc{PolySkill}'s interface typing could serve as the type system for our parameter lists, and \textsc{SkillEvo}'s fine-grained reward model could replace our binary success filter, which would relax the verifier dependence discussed in §\ref{sec:dependence}.

\end{document}